\documentclass{article}
\usepackage{ijcai17}

\usepackage{times}
\usepackage{epsfig}
\usepackage{graphicx}
\usepackage{amsmath}
\usepackage{amssymb}
\usepackage{float}

\usepackage{bm}
\usepackage{algorithm}
\usepackage{algorithmic}
\usepackage{helvet}
\usepackage{courier}
\usepackage{subfigure}
\usepackage{enumerate}
\usepackage{multirow}

\usepackage{balance} 
\usepackage[breaklinks=true,bookmarks=false]{hyperref}

\renewcommand{\algorithmicrequire}{ \textbf{Input:}} 
\renewcommand{\algorithmicensure}{ \textbf{Output:}} 

\newtheorem{MyDef}{Definition}

\title{ Supervised Deep Hashing for Hierarchical Labeled Data\thanks{Submission to AAAI 2018}}  

\author{Dan Wang$^\spadesuit$, Heyan Huang$^\spadesuit$, Chi Lu$^\spadesuit$, Bo-Si Feng$^\spadesuit$, Liqiang Nie$^\heartsuit$, Guihua Wen$^\clubsuit$, Xian-Ling Mao$^\spadesuit$\\
 $^\spadesuit$Department of Computer Science, Beijing Institute of Technology, China \\ 
  $^\heartsuit$Department of Computing, National University of Singapore, Singapore \\ 
  $^\clubsuit$Department of Computer Science and Technology, South China University of Technology, China\\ 
  {\tt \{wangdan12856, hhy63, luchi, 2120160986, maoxl\}@bit.edu.cn} \\
 {\tt nieliqiang@gmail.com, crghwen@scut.edu.cn}
}

\begin{document}  

\maketitle

\begin{abstract}
Recently, hashing methods have been widely used in large-scale image retrieval. However, most existing hashing methods did not consider the hierarchical relation of labels, which means that they ignored the rich information stored in the hierarchy. Moreover, most of previous works treat each bit in a hash code equally, which does not meet the scenario of hierarchical labeled data. In this paper, we propose a novel deep hashing method, called supervised hierarchical deep hashing (SHDH), to perform hash code learning for hierarchical labeled data.   
Specifically,   
we define a novel similarity formula for hierarchical labeled data by weighting each layer,   
and design a deep convolutional neural network to obtain a hash code for each data point. Extensive experiments on several real-world public datasets show that the proposed method outperforms the state-of-the-art baselines in the image retrieval task. 
\end{abstract}

\section{Introduction}

Due to its fast retrieval speed and low storage cost, similarity-preserving hashing has been widely used for approximate nearest neighbor (ANN) search \cite{arya1998optimal,zhu2016deep}. The central idea of hashing is to map the data points from the original feature space into binary codes in the Hamming space and preserve the pairwise similarities in the original space.    
 With the binary-code representation, hashing enables constant or sub-linear time complexity for ANN search \cite{Gong2011Iterative,Zhang2014Supervised}. Moreover, hashing can reduce the storage cost dramatically. 

Compared with traditional data-independent hashing methods like Locality Sensitive Hashing (LSH) \cite{Gionis2000Similarity} which do not use any data for training, data-dependent hashing methods,  can achieve better accuracy with shorter codes by learning hash functions from training data \cite{Gong2011Iterative,Liu2012Supervised,Liu2014Discrete,Zhang2014Supervised}.   
Existing data-dependent methods can be further divided into three categories: unsupervised methods \cite{He2013K,Gong2011Iterative,Liu2014Discrete,Shen2015Learning}, semi-supervised methods \cite{5539994,Wang2014Learning,zhang2016ssdh}, and supervised methods  \cite{Liu2012Supervised,Zhang2014Supervised,AAAI1612353}.  
Unsupervised hashing works by preserving the Euclidean similarity between the attributes of training points, while semi-supervised and supervised hashing try to preserve the semantic similarity constructed from the semantic labels of the training points  \cite{Zhang2014Supervised,norouzi2011minimal,AAAI1612353}. Although there are also some works to exploit other types of supervised information like the ranking information for hashing \cite{li2013learning,norouzi2012hamming},    
the semantic information is usually given in the form of pairwise labels indicating whether two data points are known to be similar or dissimilar. 
Meanwhile, some recent supervised methods performing simultaneous feature learning and hash code learning with deep neural networks, have shown better performance   \cite{Zhang2015Bit,Zhao2015Deep,Li2015Feature,DBLPconfijcaiLiWK16}.  
Noticeably, these semi-supervised and supervised methods can mainly be used to deal with the data with non-hierarchical labels.  

\begin{figure}[t]
\centering
\includegraphics[width=0.4\textwidth]{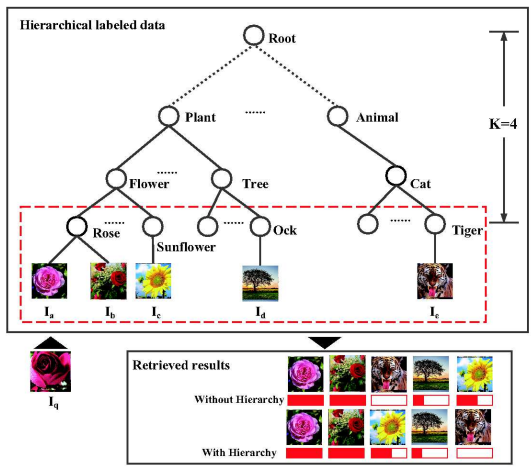}
\caption{A hierarchical labeled dataset. The height of the hierarchy is four. The different retrieved results where whether the hierarchical relation is considered are demonstrated.  
The longer a red bar is,  the more similar the corresponding image is.}  
\label{labeltree}
\end{figure}

However, there are indeed lots of hierarchical labeled data, such as Imagenet\cite{Deng2009ImageNet}, IAPRTC-12\footnote{http://imageclef.org/SIAPRdata.} and CIFAR-100\footnote{https://www.cs.toronto.edu/~kriz/cifar.html.}. Intuitively, we can simply take hierarchical labeled data as non-hierarchical labeled data, and then take advantage of the existing algorithms. Obviously, it cannot achieve optimal performance, because most of the existing methods are essentially designed to deal with non-hierarchical labeled data which do not consider special characteristics of hierarchical labeled data.
For example, in Figure \ref{labeltree}, if taking the hierarchical ones as non-hierarchical labeled data, images $I_a$ and $I_b$ have the same label ``Rose", the label of the image $I_c$ is ``Sunflower", and the labels for $I_d$ and $I_e$ are respectively ``Ock" and ``Tiger".  Given a query $I_q$ with the ground truth label ``Rose", the retrieved results may be ``$I_a$, $I_b$, $I_e$, $I_d$, and $I_c$"  in descending order without considering the hierarchy. It does not make sense that the ranking positions of images $I_e$ and $I_d$ are higher than that of $I_c$, because the image $I_c$ is also a flower although it is not a rose.  

To address the aforementioned problem, we propose a novel supervised hierarchical deep hashing method for hierarchical labeled data, denoted as SHDH.    
Specifically, we define a novel similarity formula for hierarchical labeled data by weighting each layer, and design a deep convolutional neural network to obtain a hash code for each data point. Extensive experiments on several real-world public datasets show that the proposed method outperforms the state-of-the-art baselines in the image retrieval task.

\section{Method}

\subsection{Hierarchical Similarity}  
\label{sec:hier-simil}

It is reasonable that images have distinct similarity in different layers in a hierarchy. For example, in Figure \ref{labeltree}, images $I_a$ and $I_c$ are similar in the third layer because they are both flower. However, they are dissimilar in the fourth layer because $I_a$ belongs to rose but $I_c$ belongs to sunflower. In the light of this, we have to define hierarchical similarity for two images in hierarchical labeled data.  

\begin{MyDef}[Layer Similarity] For two images $i$ and $j$, the similarity at the $k^{th}$ layer in the hierarchy is defined as: 
\begin{equation}
s^k_{ij}=\left\{ \begin{array}{ll}
1, & \textrm{if  $ Ancestor_k(i) = Ancestor_k(j)$, $k \neq 1$; }\\
0, & \textrm{otherwise.}
\end{array} \right.
\label{simk}
\end{equation}
where $Ancestor_k(i)$ is the ancestor node of image $i$ at the $k^{th}$ layer.     
\end{MyDef}
Equation (\ref{simk}) means that if images $i$ and $j$ share the common ancestor node in the $k^{th}$ layer, they are similar at this layer. On the contrary, they are dissimilar. For example, in Figure \ref{labeltree}, the layer similarities between images $I_a$ and $I_c$ at different layers are: $s^1_{I_aI_c} = 0$, $s^2_{I_aI_c} = 1$, and $s^4_{I_aI_c} = 0$.  




Intuitively, the higher layer is more important, because we cannot reach the right descendant nodes if we choose a wrong ancestor. We thus have to consider the weight for each layer in a hierarchy. 
\begin{MyDef}[Layer Weight] 
The importance of $k^{th}$ layer in a hierarchy whose height is K, can be estimated as:   
\begin{equation}
\label{eq:layerweight}
u_k = \frac{2(K+1-k)}{K(K-1)},
\end{equation}
where $k \in [ 2,3, \cdots,K]$. 
\end{MyDef}
Note that $u_1 = 0$ because the root has no discriminative ability for all data points. It is easy to prove that $u_{k} > u_{k+1}$, which satisfies the demand where the influence of ancestor nodes is greater than that of descendant nodes. And $\sum_{k=0}^Ku_k=1$. 

Based upon the two definitions above, the final hierarchical similarity between images $i$ and $j$ can be calculated as below:
\begin{equation}  
    s_{ij}=2\sum^K_{k=1}u_ks^k_{ij}-1,
\label{sim}
\end{equation}
where $K$ is the height of a hierarchy. 
Equation (\ref{sim}) guarantees that the more common hierarchical labels image pairs have, the more similar they are.  
  

\begin{figure*}[htbp]
\centering
\includegraphics[width=0.85\textwidth]{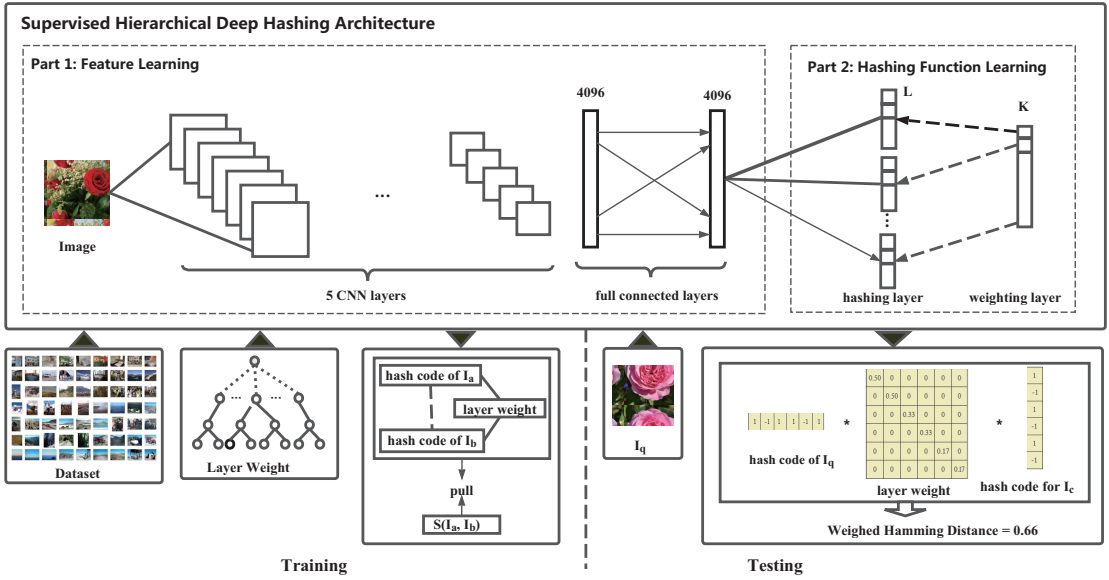}
\caption{The SHDH learning framework.  
It takes raw images as input. The training stage is illustrated in the left bottom corner. A retrieval example in testing stage is presented in the right bottom corner.}
\label{hierarmodel}
\end{figure*}

\subsection{Supervised Hierarchical Deep Hashing} 
\label{model}
Figure \ref{hierarmodel} shows the deep learning architecture of the proposed method. 
Our SHDH model consists of two parts: feature learning and hash function learning. The feature learning part includes a convolutional neural network (CNN) component and two fully-connected  layers. The CNN component contains five convolutional layers.     
After the CNN component, the architecture holds two fully-connected layers which have 4,096 units. The mature neural network architecture in \cite{Chatfield2014Return} is multipled in our work. Other CNN architectures can be used here such as AlexNet \cite{Krizhevsky2012ImageNet}, GoogleNet \cite{SzegedyLJSRAEVR15}. Since our work fouces on the influence of hierarchy, the CNN part is not the key point. The activation function used in this part is Rectified Linear Units (ReLu). More details can be referred in \cite{Chatfield2014Return}. 

The hash function learning part includes a hashing layer and an independent weighting layer. 
The hashing functions are learnt by the hashing layer whose size is the length of hash codes. And no activation function used in this layer. 
Note that the hashing layer is divided into $K$-segments and $K$ is the height of hierarchical labeled data. The size of $1^{st}$ $\sim$ $(K-1)^{th}$ segments is $\left \lfloor \frac{L}{K} \right \rfloor$, where L is the length of hash codes.  And the size of the $K^{th}$ segment is $L - \left \lfloor \frac{L}{K} \right \rfloor \times (K-1) $. The size of $k^{th}$ segment is denoted as $L_k$. Here, there is an implicit assumption that $L$ is larger than $K$. 
Besides, the values in the weighting layer are the weights learnt by Eq. (\ref{eq:layerweight}) from the hierarchical labeled data, which are used to adjust the Hamming distance among hash codes.  Each value in the weighting layer weights a corresponding segment in the hashing layer.
The parameters including weights and bias in the hashing layer are initialized to be a samll real number between 0 and 0.001. 


\subsubsection{Objective Function}


Given a hierarchical labeled dataset $\bm{X}=\{\bm{x_i} \}_{i=1}^N$ where $\bm{x_i}$ is the feature vector for data point $i$ and $N$ is the number of data points. Its semantic matrix $\bm{S}=\{s_{ij}\}$ can be built via Eq. (\ref{sim}), where $s_{ij} \in [-1,1]$.   
The goal of our SHDH is to learn a $L$-bit binary codes vector $\bm{h_i} = [h_{i1}, ..., h_{iL}] \in\{-1,1\}^{L\times 1}$ for each point $\bm{x_i}$.  
Assume there are $M + 1$ layers in our deep network, and there are $e^m$ units for the $m^{th}$ layer, where $m = 1, 2, ..., M$.  
For a given sample $\bm{x_i} \in \mathbb{R}^{d}$, the output of the first layer is: $\bm{\phi}_i^1 = f(\bm{W}^1\bm{x_i} + \bm{v}^1) \in \mathbb{R}^{e^1}$, where $\bm{W}^1$ $\in \mathbb{R}^{e^1\times d}$ is the projection matrix to be learnt at the first layer of the network, $\bm{v}^1 \in \mathbb{R}^{e^1}$ is the bias, and $f(\cdot)$ is the activation function.   
The output of the first layer is then considered as the input for the second layer, so that $\bm{\phi}^2_i = f(\bm{W}^2\bm{\phi}^1_i + \bm{v}^2) \in \mathbb{R}^{e^2}$, where $\bm{W}^2 \in \mathbb{R}^{e^2 \times e^1}$ and $\bm{v}^2 \in \mathbb{R}^{e^2}$ are the projection matrix and bias vector for the second layer, respectively. Similarly, the output for the $m^{th}$ layer is: $\bm{\phi}_n^m = f(\bm{W}^m\bm{\phi}_i^{m-1} +  \bm{v}^m)$, and the output at the top layer of our network is:  
\[
g(\bm{x_i}) = \bm{\phi}_i^M = f(\bm{W}^M\bm{\phi}_i^{M-1} + \bm{v}^M),
\]
where the mapping $g:\mathbb{R}^d \rightarrow \mathbb{R}^{e^M}$ is parameterized by $\{\bm{W}^m, \bm{v}^m\}_{m=1}^M$, $1 \le m \le M$. 
Now, we can perform hashing for the output $h^M$ at the top layer of the network to obtain binary codes as follows: $\bm{h_i} = sgn(\bm{\phi}_i^M)$. The procedure above is forward. To learn the parameters of our network, we have to define an objective function.   


First, for an image $\bm{x_i}$, its hash code is $\bm{h(i)}\in  \{-1,1\}^{L \times 1}$ consisting of $\bm{h^k(i)}$, where $\bm{h^k(i)}$ is the hash code in the $k^{th}$ segment, $k \in \{1, ..., K\}$. Thus, the {\bf weighted Hamming distance} between images $i$ and $j$ can be defined as:  
\begin{equation}
D(\bm{h(i),h(j)}) = \sum^K_{k=1}u_k\bm{h^k(i)}^T\bm{h^k(j))}  .
\end{equation}
We define the similarity-preserving objective function:  
\begin{equation}
E_1= \min\sum^K_{k=1}(\frac{1}{L_k}u_k\bm{h^k(i)}^T\bm{h^k(j))}-s^k_{ij})^2
\label{eq1}
\end{equation}
Eq. (\ref{eq1}) is used to make sure the similar images could share same hash code in each  
segment.  

Second, to maximize the information from each hash bit, each bit should be a balanced partition of the dataset.  Thus,  we maximize the entropy, just as below:   
\begin{equation}
E_2 = \max \sum^K_{k=1}u_k\bm{h^k(i)}^T\bm{h^k(j))}.
\label{eq2}
\end{equation}

We ultimately combine Eq. (\ref{eq1}) and Eq. (\ref{eq2}) to obtain the total objective function:   
\begin{eqnarray}
\label{eq:obj}
J =&\min \sum^K_{k=1}(\frac{1}{L_k}u_k\bm{h^k(i)}^T\bm{h^k(j))}-s^k_{ij})^2   \nonumber  \\ 
& +  \alpha \max \sum^K_{k=1}u_k\bm{h^k(i)}^T\bm{h^k(j))},  \nonumber  \\  
\end{eqnarray}
where $\alpha$ is hyper-parameter. 

\subsubsection{Learning}
Assume that $\bm{H}=[\bm{h_1}^T; \bm{h_2}^T; ...; \bm{h_N}^T]$ is all the hash codes for $N$ data points, and thus the objective function Eq. (\ref{eq:obj}) could be transformed into the matrix form as below:  
\begin{equation}
J  = \min \|\bm{HAH}^T-L\bm{S}\|_F^2 -\alpha tr(\bm{HAH}^T).
\label{equJ}
\end{equation}
$\bm{H} = sgn(\bm{W}^m\bm{\phi}^m+\bm{v}^m)$, where $sgn(\centerdot)$ denotes the elementwise sign function which returns $1$ if the element is positive and returns $-1$ otherwise;   $\bm{W}^m$ is the weight of the $m^{th}$ layer in our SHDH, $\bm{v}^m$ is bias vector, and $\bm{\phi} ^m $ is the output of the $m^{th}$ layer. $\bm{A}\in \mathbb{R}^{L \times L}$ is a diagonal matrix. It can be divided into $K$ small diagonal matrix corresponding to $L_k$. The diagonal value of $A_k$ is $u_k$.  
Since the elements in $\bm{H}$ are discrete integer, $J$ is not derivable. So, we relax it as $ \bm{\tilde H}$ from discrete to continuous by removing the sign function.  
Stochastic gradient descent (SGD) \cite{Paras2014Stochastic} is used to learn the parameters. We use back-propagation (BP) to update the parameters:  

\begin{equation}
\label{equwv}
\left\{ \begin{array}{ll}
\frac{\partial J}{\partial \bm{W}^m} &= \frac{\partial J}{\partial \bm{Z^m}}\cdot \frac{\partial \bm{Z^m}}{\partial \bm{W}^m} = \bigtriangleup\cdot (\bm{\phi} ^m)^T,\\
\frac{\partial J}{\partial \bm{v}^m} &=\bigtriangleup ^m,
\end{array} \right.
\end{equation}

where $\bigtriangleup$ is calculated as below:  
\begin{equation}
\begin{split}
\small
\bigtriangleup ^m =&\frac{\partial J}{\partial\bm{\tilde H}}\cdot \frac{\partial\bm{\tilde H}}{\partial \bm{Z^m}} \\
= &\{2(\bm{\tilde H}^m\bm{A}^T(\bm{\tilde H}^m)^T\bm{\tilde H}^m\bm{A}+\bm{\tilde H}^m\bm{A}(\bm{\tilde H}^m)^T\bm{\tilde H}^m\bm{A}^T  \nonumber \\
&-\bm{S}^T\bm{\tilde H}^m\bm{A} - \bm{S\tilde H}^m\bm{A}^T)-\alpha\bm{\tilde H}^m\bm{A}^T-\alpha\bm{\tilde H}^m\bm{A}\}  \nonumber \\
&\odot f'(\bm{W}^m\bm{\tilde H}^{m-1} + \bm{v}^m). \nonumber 
\end{split}
\end{equation}
where $\odot$ denotes element-wise multiplication. 

Then, the parameters are updated by using the following gradient descent algorithm until convergence:   
\begin{equation}
\label{equww}
\bm{W}^m = \bm{W}^m -\eta\frac{\partial J}{\partial \bm{W}^m},
\end{equation}
\begin{equation}
\label{equvv}
\bm{v}^m = \bm{v}^m -\eta\frac{\partial J}{\partial \bm{v}^m}, 
\end{equation}
where $\eta$ is the step-size.   
In addition, we use the lookup table technology proposed in \cite{Zhang2015Bit} to speed up the searching process.  
The outline of the proposed supervised hierarchical deep hashing (SHDH) is described in Algorithm \ref{alg}.

\begin{algorithm}[t]
\caption{The Learning Algorithm for SHDH}
\label{alg}  
\begin{algorithmic}[1]
\REQUIRE ~~
{Training images $\bm{X}=\{\bm{x_i} \}_{i=1}^N$, the length of hash code $L$, the height of label tree $K$, the max iterative count $T$, the size of minibatch $M$ (default 128), the learning rate $\eta$ (default $0.01$)}
\ENSURE ~~
{The network parameters $\{\bm{W},\bm{v}\}$}
\STATE {Initialize net weights and bias for hashing layer}
\STATE $\bm{S} \gets $ using Eq. (\ref{sim}), $\bm{S} \in \mathbb{R}^{N \times N}$
\REPEAT
\STATE  Update $\eta$ using $\frac{2}{3}\times \eta$ every 20 iterations.
\STATE  Randomly sample from $\bm{X}$ to get a minibatch. \\
For each image $\bm{x_i}$, perform as below: 
 \FOR{$k=1,\cdots,K$}
 \STATE   Calculate the output $\bm{h^k(x_i)}$ of image $\bm{x_i}$ by forward propagation. 
\ENDFOR
\STATE {Merge $\{\bm{h^k(x_i)}\}_{k=1}^K$, to get $\bm{H}$}
 \STATE   Calculate $\frac{\partial J}{\partial \bm{W}^m}$,$\frac{\partial J}{\partial \bm{v}^m}$ according to Eq. (\ref{equwv}).
 \STATE   Update the parameters $\{\bm{W}^m,\bm{v}^m\}$ by back propagation according to Eq. (\ref{equww}) and (\ref{equvv}). 
\UNTIL{up to $T$}
\end{algorithmic}
\end{algorithm}

\begin{table*}[h]  
\centering
\caption{Results on the CIFAR-100 dataset. The ranking results are measured by ACG, DCG, and NDCG@N (N=100). }
\begin{tabular}{|c|ccc|ccc|ccc|}
\hline
\multicolumn{10}{|c|}{CIFAR-100}                                                                                                                                                                                                                                                                                            \\ \hline
\multicolumn{1}{|c|}{\multirow{2}{*}{Methods}} & \multicolumn{3}{c|}{ACG@100}                                                & \multicolumn{3}{c|}{DCG@100}                                                                     & \multicolumn{3}{c|}{NDCG@100}                                                           \\\cline{2-10}
\multicolumn{1}{|c|}{}                         & 32     & 48     & 64              & 32               & 48              & 64               & 32     & 48    & 64    \\ \hline\hline
KMH         & 0.2023 & -      & 0.2261          & 6.0749                                & -                                     & 6.7295           & 0.4169 & -      & 0.4189 \\ 
ITQ         & 0.2091 & 0.2312 & 0.2427          & 6.1814                                & 6.7583                                & 7.0593           & 0.4197 & 0.4243 & 0.4272 \\
COSDISH+H   & 0.1345 & 0.1860 & 0.2008          & 4.2678                                & 5.5619                                & 5.9169           & 0.4072 & 0.4417 & 0.4523 \\ 
KSH+H       & 0.1611 & 0.1576 & 0.1718          & 4.9904                                & 4.9282                                & 5.3378           & 0.3940 & 0.3897 & 0.3924 \\ 
DPSH        & 0.4643                      & 0.4973                      & 0.5140          & 11.5129          & 12.2878          & 12.7072          & 0.5650                      & 0.5693                      & 0.5751                      \\ 
COSDISH     & 0.1366                      & 0.1428                      & 0.1501          & 4.5079           & 4.6957           & 4.8601           & 0.4063                      & 0.4156                      & 0.4127                      \\ 
LFH         & 0.1152                      & 0.1291                      & 0.1271          & 3.7847           & 4.3299           & 4.3239           & 0.3924                      & 0.4008                      & 0.4011                      \\ 
KSH         & 0.1291                      & 0.1393                      & 0.1509          & 3.3520           & 4.3009           & 4.8293           & 0.3711                      & 0.3766                      & 0.3763                      \\  
SHDH        & \textbf{0.5225}             & \textbf{0.5724}             & \textbf{0.6084} & \textbf{12.7460} & \textbf{13.9575} & \textbf{14.7861} & \textbf{0.6141}             & \textbf{0.6281}             & \textbf{0.6406}             \\ \hline
\end{tabular}
\label{cifar}
\end{table*}

\section{Experiments}

\subsection{Datasets and Setting} 
We carried out experiments on two public benchmark datasets: CIFAR-100 dataset and IAPRTC-12 dataset.
CIFAR-100 is an image dataset containing 60,000 colour images of 32$\times$32 pixels. It has 100 classes and each class contains 600 images. The 100 classes in the CIFAR-100 are grouped into 20 superclasses. Each image has a ``fine'' label (the class which it belongs to) and a ``coarse'' label (the superclass which it belongs to). Thus, the height of the hierarchical labels with a ``root'' node in CIFAR-100 is three.  
 The IAPRTC-12 dataset has 20,000 segmented images. Each image has been manually segmented, and the resultant regions have been annotated according to a predefined vocabulary of labels. The vocabulary is organized according to a hierarchy of concepts. The height of the hierarchical labels in IAPRTC-12 is seven. For both datasets, we randomly selected 90\% as the training set and the left 10\% as the test set.   
The hyper-parameter $\alpha$ in SHDH is empirically set as one.

\begin{figure*}[htb!]
\centering
\includegraphics[width=0.98\textwidth]{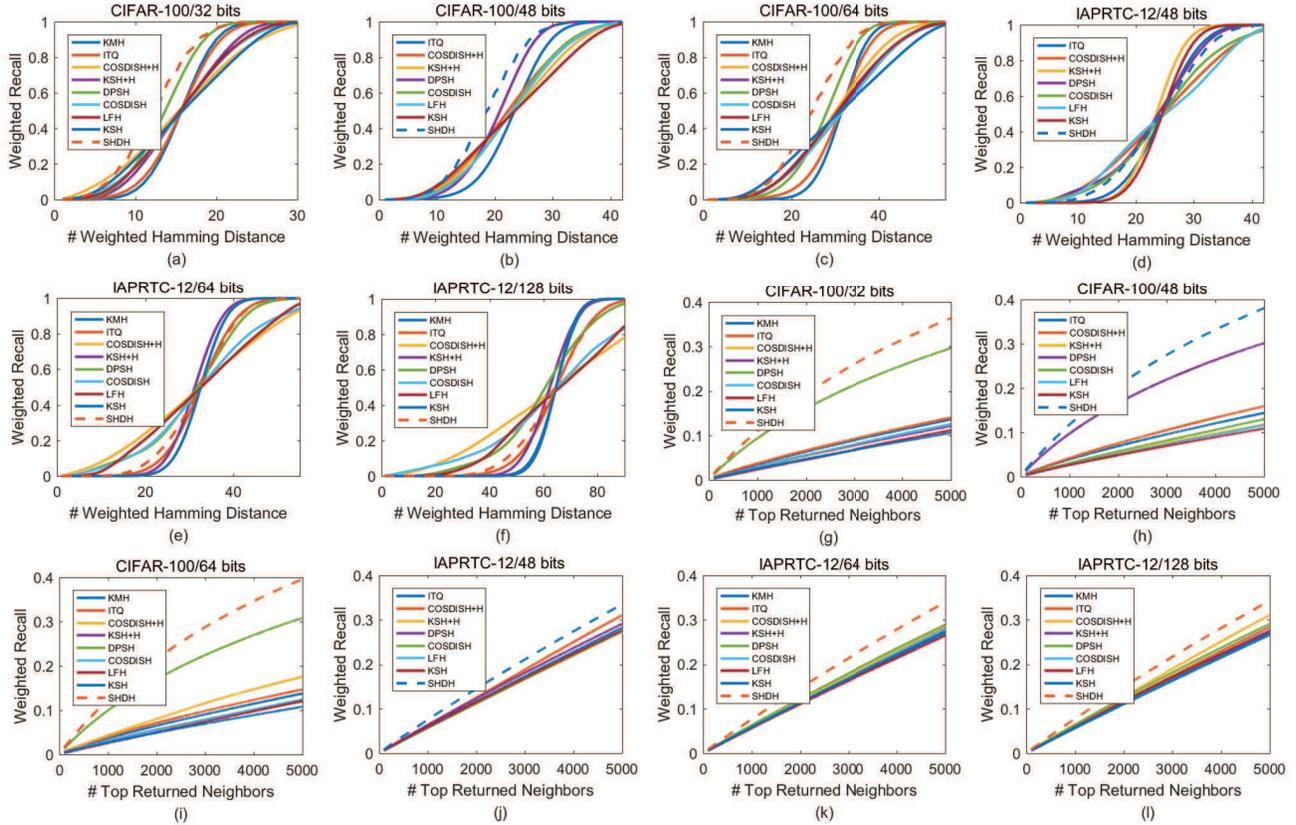}
\caption{Weighted Recall curves on CIFAR-100 and  IAPRTC-12. (a) $\sim$ (f) show the Weighted Recall within various weighted Hamming distance at different number of bits. (g) $\sim$ (l) show the Weighted Recall@n at different number of bits.} 
\label{pic}
\end{figure*}   



We compared our methods with six state-of-the-art hashing methods, where four of them are supervised, the other two are unsupervised. 
The four supervised methods include  DPSH \cite{Li2015Feature,DBLPconfijcaiLiWK16}, COSDISH \cite{AAAI1612353}, LFH \cite{Zhang2014Supervised}, and KSH \cite{Liu2012Supervised}. The two unsupervised methods are KMH \cite{He2013K} and ITQ \cite{Gong2011Iterative}.   
For all of these six baselines, we employed the implementations provided by the original authors, and used the default parameters recommended by the corresponding papers.    
Moreover, to study the influence of hierarchical labels separately, we replaced the values in the similarity matrix for KSH and COSDISH by using hierarchical similarity  
to obtain two new methods, KSH+H and COSDISH+H. ``H'' means hierarchical version. ITQ and KMH cannot be modified because they are unsupervised. LFH and DPSH cannot be modified because their algorithm structures are not suitable to add hierarchical labeled information.

 We resized all images to 32$\times$32 pixels and directly used the raw images as input for the deep hashing methods including SHDH and DPSH. The left six methods use hand-crafted features. We represented each image in CIFAR-100 and IAPRTC-12 by a 512-D GIST vector.    

\subsection{Evaluation Criterion}
We measured the ranking quality of retrieved list for different methods by Average Cumulative Gain (ACG), Discounted Cumulative Gain (DCG), Normalized Discounted Cumulative Gain (NDCG) \cite{Rvelin2000IR} and Weighted Recall. Note that we proposed the Weighted Recall metric to measure the recall in the scenario of hierarchical labeled data, defined as:   
\begin{equation*}
Weighted \  Recall(q)@n=\frac{\sum^n_{i=1}s_{qi}}{\sum^N_{i=1}s_{qi}},
\end{equation*}
where $n$ is the number of top returned data points, $s_{qi}$ represents the similarity between the query $q$ and $i^{th}$ data point in the ranking list, $N$ is the length of the ranking list.

\begin{table*}[htb!]
\centering
\caption{Results on the IAPRTC-12 dataset. The ranking results are evaluated by ACG, DCG, and NDCG@N (N=100).} 
\begin{tabular}{|c|ccc|ccc|ccc|}
\hline
\multicolumn{10}{|c|}{IAPRTC-12}                                                                                                                                                                                                                                                                                           \\ \hline
\multicolumn{1}{|c|}{\multirow{2}{*}{Methods}} & \multicolumn{3}{c|}{ACG@100}                                                & \multicolumn{3}{c|}{DCG@100}                                                                     & \multicolumn{3}{c|}{NDCG@100}                                                           \\ \cline{2-10}
\multicolumn{1}{|c|}{}                        &  48     & 64     & 128              & 48               & 64              & 128               & 48     & 64    & 128    \\ \hline\hline
    KMH     & -        & 3.7716  & 3.7446  & -        & 87.5121 & 87.0493 & -        & 0.6427   & 0.6373  \\ 
                                  ITQ     & 3.8351   & 3.8502  & 3.8609  & 88.5562  & 88.9057 & 89.2016 & 0.6626   & 0.6633   & 0.6652 \\
COSDISH+H & 3.8249   & 3.7245  & 3.8448  & 88.3121  & 86.3037 & 88.5056 & 0.6957   & 0.6885   & 0.6970 \\ 
KSH+H     & 3.7304  & 3.7535  & 3.7779   & 86.5606  & 87.0894 & 87.5743 & 0.6459   & 0.6494   & 0.6518 \\
DPSH    & 4.0085   & 4.0227  & 4.0980  & 91.4972  & 92.0570 & 93.4613 & 0.6618   & 0.6607   & 0.6630  \\
COSDISH & 3.6856   & 3.6781  & 3.7018  & 85.2368  & 85.1622 & 85.7606 & 0.6412   & 0.6443   & 0.6408  \\
LFH     & 3.7076   & 3.6851  & 3.6988  & 85.7599  & 85.2662 & 85.6601 & 0.6390   & 0.6365   & 0.6400  \\
KSH     & 3.8357   & 3.8317  & 3.7909  & 88.5041  & 88.5589 & 87.8282 & 0.6507   & 0.6482   & 0.6408  \\
SHDH        &\textbf{ 4.4870}   &\textbf{ 4.5284}  &\textbf{ 4.5869}  & \textbf{100.6373} & \textbf{101.4812} & \textbf{102.6919} & \textbf{0.7372}   &\textbf{ 0.7440}   & \textbf{0.7489}  \\\hline
\end{tabular}
\label{saia}
\end{table*}
  
\subsection{Results on CIFAR-100}
Table \ref{cifar} summarizes the comparative results of different hashing methods on the CIFAR-100 dataset.   
We have several observations from Table \ref{cifar}:  
(1) our SHDH outperforms the other supervised and unsupervised baselines for different code length. For example, comparing with the best competitor (DPSH), the results of our SHDH have a relative increase of 12.5\% $\sim$ 18.4\% on ACG,  10.7\% $\sim$ 16.7\% on DCG, and 8.7\% $\sim$ 11.4\% on NDCG;  
(2) the hierarchical semantic labels can improve the performance of hashing methods. For example, COSDISH+H and KSH+H perform respectively better than COSDISH and KSH,   
which means the inherent hierarchical information is valuable to improve hashing performance;  (3) among all the supervised approaches, the deep learning based approaches (SHDH and DPSH) give relatively better results, and it confirms that the learnt representations by deep network from raw images are more effective than hand-crafted features to learn hash codes. 

\begin{figure}[tb!]
\centering
\includegraphics[width=1.0\linewidth]{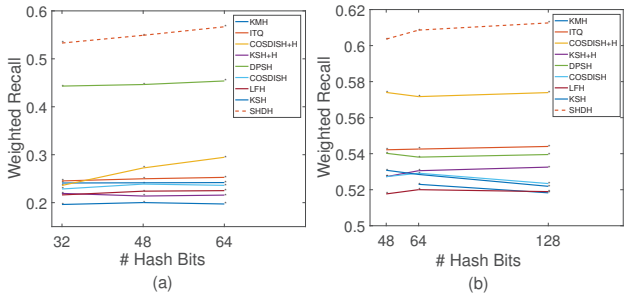}
\caption{Weighted Recall@n (n=10,000) over (a) CIFAR-100 and (b) IAPRTC-12. }
\label{figrecalltopK}
\end{figure}

Figure \ref{pic} (a) $\sim$ (c) are the Weighted Recall curves for different methods over different weighted Hamming distance at 32, 48, and 64 bits, respectively, which shows our method has a consistent advantage over baselines.  
Figure \ref{pic} (g) $\sim$ (i) are the Weighted Recall results over top-$n$ retrieved results, where $n$ ranges from 1 to 5,000. Our approach also outperforms other state-of-the-art hashing methods. The Weighted Recall curves at different length of hash codes are also illustrated in Figure \ref{figrecalltopK} (a). From the figure, our SHDH model performs better than baselines, especially when the code length increases. This is because when the code length increases, the learnt hash functions can increase the discriminative ability for hierarchical similarity among images.   

\subsection{Results on IAPRTC-12}
 Table \ref{saia} shows the performance comparison of different hashing methods over IAPRTC-12 dataset, and our SHDH performs better than other approaches regardless of the length of codes. 
Obviously, it can be found that all baselines cannot achieve optimal performance for hierarchical labeled data. 
Figure \ref{pic} (j) $\sim$ (l) are the Weighted Recall results over top-$n$ returned neighbors, where $n$ ranges from 1 to 5,000. These curves show a consistent advantage against baselines. Moreover,  our SHDH provides the best performance at different code length, shown in Figure \ref{figrecalltopK} (b). 

The results of the Weighted Recall over different weighted Hamming distance are shown in Figure \ref{pic} (d) $\sim$ (f). In these figures, our method is not the best one. The reason is that our SHDH has better discriminative ability at the same weighted Hamming distance due to considering the hierarchical relation. For example, DPSH returns 4,483 data points while our SHDH only returns 2,065 points when the weighted Hamming distance is zero and the code length is 64 bits. Thus, the better discriminative ability leads to better precision (Table \ref{saia}) but not-so-good Weighted Recall.   


\subsection{Sensitivity to Hyper-Parameter}
\begin{figure}[!tb]
\centering
\includegraphics[width=0.35\textwidth]{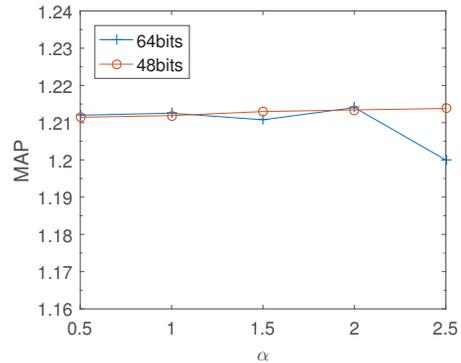}
\caption{Sensitivity to hyper-parameter $\alpha$ over CIFAR-100.}
\label{sens-alpha} 
\end{figure}

Figure \ref{sens-alpha} shows the effect of the hyper-parameter $\alpha$ over CIFAR-100. We can find that SHDH is not sensitive to $\alpha$. For example, SHDH can achieve good performance on both datasets with 0.5 $\le$ $\alpha$ $\le$ 2. We can also obtain similar conclusion over IAPRTC-12 dataset, and the figure is not included in this paper due to the limitation of space.  
  
\section{Conclusion}
In this paper, we have proposed a novel supervised hierarchical deep hashing method for hierarchical labeled data. 
To the best of our knowledge, SHDH is the first method to  utilize the hierarchical labels of images in supervised hashing area. 
Extensive experiments on two real-world public datasets have shown that the proposed SHDH method outperforms the state-of-the-art hashing algorithms. 

In the future, we will explore more hashing methods to process hierarchical labeled data, and further improve the performance of hashing methods for non-hierarchical labeled data by constructing their hierarchy automatically.




\bibliographystyle{named}
\bibliography{ijcai17}

\end{document}